\documentclass[11pt]{article}

\usepackage[preprint]{acl}

\usepackage{times}
\usepackage{latexsym}

\usepackage[T1]{fontenc}

\usepackage[utf8]{inputenc}

\usepackage{microtype}

\usepackage{inconsolata}

\usepackage{graphicx}
\usepackage{bbm}

\usepackage{svg}
\usepackage{graphicx}
\usepackage{fontawesome5}
\usepackage{hyperref}
\usepackage{booktabs}
\usepackage{custom_commands}
\usepackage{multirow}
\usepackage[table,xcdraw]{xcolor}
\definecolor{webblue}{RGB}{93, 169, 221}
\usepackage{lscape}
\RequirePackage{algorithm}
\RequirePackage{algorithmic}
\usepackage[capitalize,noabbrev]{cleveref}
\usepackage{subcaption}
\usepackage{wrapfig}

\usepackage[utf8]{inputenc}
\usepackage[most]{tcolorbox}
\usepackage{listings}
\usepackage{enumitem}
\setlist[enumerate]{itemsep=0mm}


\usepackage{pifont}

\usepackage{tcolorbox}
\tcbuselibrary{breakable,skins}
\usepackage{dblfloatfix} 
\tcbset{
  guiPrompt/.style={
    colback=gray!10,            
    colframe=black,             
    coltitle=white,             
    colbacktitle=gray!30!black, 
    fonttitle=\bfseries,
    fontupper=\ttfamily,        
    boxrule=0.5pt,
    arc=2mm,                    
    top=0.6em, bottom=0.6em, left=0.8em, right=0.8em,
    enhanced,
    breakable
  }
}
\newif\ifcomments
\commentstrue            

\newcommand{\cmark}{\textcolor{green!60!black}{\ding{51}}} 
\newcommand{\xmark}{\textcolor{red}{\ding{55}}} 
\newcommand{\webstar}{\texttt{WebSTAR}}
\newcommand{\webscore}{\texttt{WebSCORE}}
\newcommand{\steprm}{\texttt{StepRM}}

\title{WebSTAR: Scalable Data Synthesis for Computer Use Agents with Step-Level Filtering}

\author{\textbf{Yifei He\thanks{Work done during an internship at Microsoft. Correspondence email: \texttt{yifeihe3@illinois.edu}.}$^\dagger$\textsuperscript{1}, Pranit Chawla$^\dagger$\textsuperscript{2}, Yaser Souri\textsuperscript{2}, Subhojit Som\textsuperscript{2}, Xia Song\textsuperscript{2}}
\\
 \textsuperscript{1}University of Illinois Urbana-Champaign
 \textsuperscript{2}Microsoft \\
 [0.5em]
 \makebox[\textwidth][c]{
    \href{https://yifei-he.github.io/webstar-website/}{%
        \raisebox{-0.1\height}{\faGlobe} 
        \texttt{\textcolor{webblue}{Website}}%
    }
    \quad 
    \href{https://huggingface.co/datasets/microsoft/WebSTAR}{%
    \raisebox{-0.25\height}{\includegraphics[height=1.4em]{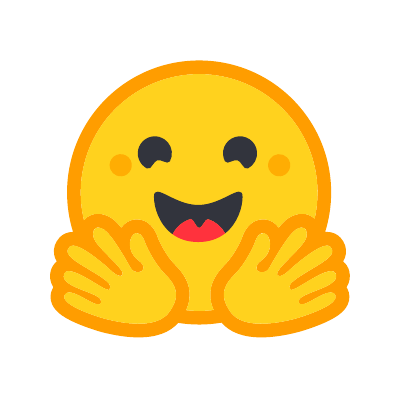}} 
    \texttt{\textcolor{webblue}{Dataset}}%
}
    \quad  
    \href{https://github.com/yifei-he/WebSTAR}{%
        \raisebox{-0.1\height}{\faGithub} 
        \texttt{\textcolor{webblue}{Code}}%
    }
  }
 }

\begin{document}
\maketitle
\def\thefootnote{$\dagger$}\footnotetext{Equal contribution}\def\thefootnote{\arabic{footnote}}

\begin{abstract}
Computer use agents (CUAs) can operate real-world digital interfaces but remain difficult to train due to the high cost of graphical user interface (GUI) interaction and the scarcity of high-quality trajectory data. Existing datasets rely on human demonstrations, limiting scalability. A natural alternative is to synthesize data from strong CUAs, yet their rollouts are highly noisy, with incorrect or suboptimal actions consisting a large proportion of the steps, making naive imitation ineffective. To tackle this challenge, we introduce a scalable data synthesis pipeline that transforms noisy rollouts into reliable supervision without human annotation. The core idea is step-level filtering, which evaluates actions individually to retain only correct steps, complemented by reasoning augmentation for improved planning. Using this pipeline, we construct \webstar, a dataset of 13.3K trajectories and 267K graded, reasoning-rich steps synthesized from OpenAI's computer-use-preview model. We train Qwen-2.5-VL-Instruct models (7B and 32B) on \webstar. On WebVoyager, our 7B model surpasses SoTA open-source CUA model UI-TARS-1.5-7B by more than 15\% with only supervised finetuning. Building on step-level grading, we further create \webscore, a dataset of graded step-level actions, and train \steprm, a 7B multimodal process reward model distilled from \texttt{o4-mini}, which matches its grading quality while being far more efficient to deploy at scale. Our results establish step-level filtering as a key principle for scalable CUA training, and we introduce two new datasets (\webstar, \webscore) along with a lightweight process reward model (\steprm) as practical tools to advance robust and efficient CUAs. 
\end{abstract} 

\section{Introduction}
Autonomous agents are increasingly capable of interpreting task instructions, perceiving environments, reasoning and planning, and executing precise actions. Among them, computer use agents (CUAs)~\citep{operator,Claude-computer-use,agasheagent,qin2025ui,wang2025ui,wang2025opencua} have emerged as a powerful class of models that can operate real-world digital interfaces such as web browsers~\citep{deng2023mind2web,he2024webvoyager,xue2025illusion} and desktop environments~\citep{xie2024osworld,bonattiwindows,wu2025gui}. Despite recent progress, training CUAs remains fundamentally challenging due to two key issues: the high cost of interaction with GUI environments, and the scarcity of high-quality training data. 

\begin{figure*}[t!]
    \centering
    \includegraphics[width=0.85\linewidth]{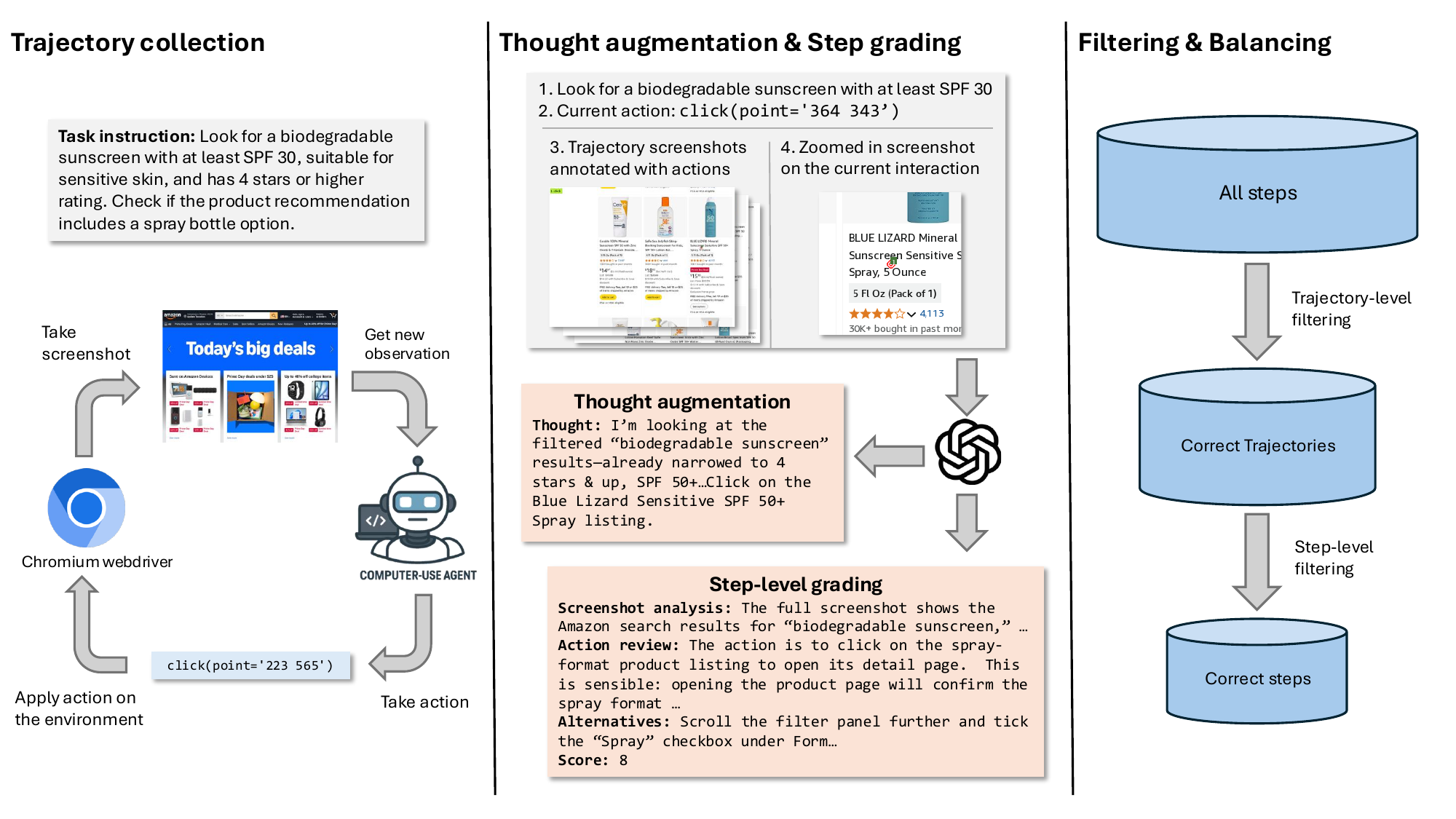}
    \caption{\textbf{Overview of our data synthesis pipeline.} Our approach consists of three main stages: \textbf{(i) Trajectory collection:} We begin by rolling out a teacher CUA in a Chromium environment, executing actions based on user instructions and capturing observations and screenshots. \textbf{(ii) Thought augmentation \& Step grading:} For each step, the action and screenshots trajectories are passed through a model to generate an intermediate thought to guide the action. A grading model also receives the same input to evaluate the current step, assigning a score from 0 to 10. \textbf{(iii) Step-level Filtering:} We retain only high-scoring steps to ensure that the agent only learns from high-quality actions. \looseness=-1}
    \label{fig:main}
    \vspace{-0.2cm}
\end{figure*}

Live interaction with GUIs typically incurs high infrastructure costs (e.g., virtual machines or cloud environments~\citep{bonattiwindows}), which limits the scale at which interaction data can be collected. As a result, most CUA training is conducted \textit{offline}, where models are trained on pre-collected trajectories using supervised finetuning (SFT)~\citep{xuaguvis,qin2025ui}. However, \textit{offline CUA trajectory data is extremely limited}. Existing data sources are either (i) manually labeled~\citep{deng2023mind2web,xuaguvis}, incurring high labor costs, or (ii) curated using general-purpose vision-language models (VLMs) such as GPT-4o~\citep{he2024webvoyager,pahuja2025explorer}, which are not grounded in agentic tasks and often rely on auxiliary information such as accessibility trees that are not available in general deployment settings.

A promising alternative is to synthesize trajectory data using strong CUAs themselves, such as OpenAI CUA~\citep{operator} or Claude Computer Use~\citep{Claude-computer-use}. This approach resembles hard distillation in language model training, where a teacher model generates outputs and a student model learns to imitate them. However, unlike textual LLM outputs, \textit{CUA trajectories are inherently noisy}. We observe that even state-of-the-art CUAs produce incorrect or suboptimal actions in more than half of the steps. Training directly on such noisy rollouts leads to degraded performance in student models, underscoring the need for more principled filtering and augmentation strategies to transform raw rollouts into high-quality training data.  \looseness=-1

In this work, we aim to answer the research question: \textit{Given a set of task instructions and access to a capable CUA, how to most effectively synthesize high-quality training data to improve agent performance?} We propose a scalable three-step pipeline for generating high-quality CUA training data from noisy teacher rollouts (illustrated in \Cref{fig:main}): 
\begin{enumerate}
    \item \textbf{Trajectory collection:} We first roll out a teacher CUA on live websites, executing actions based on user instructions and recording screenshots. This process provides the raw but noisy rollouts that serve as the basis for data synthesis.
    \item \textbf{Thought augmentation:} Proprietary CUAs do not produce explicit reasoning traces. Thus, inspired by prior work on reasoning-augmented agents~\citep{yaoreact,xuaguvis,yang2024react}, we insert an intermediate thought before each action to strengthen the agent’s reasoning and planning abilities. These thoughts provide interpretable, step-by-step guidance that grounds the action in a structured decision-making process. 
    \item \textbf{Step-level filtering:} Instead of accepting full trajectories, we apply a grading model to evaluate each action individually, retaining only those that are contextually correct and likely to contribute to task success. This ensures that the agent is only trained on high-quality steps, and we empirically find step-level filtering crucial to the agent performance. \looseness=-1
\end{enumerate}
Together, these steps allow us to transform noisy rollouts into clean, diverse, and reasoning-rich training data, enabling more reliable offline training of computer use agents.

To evaluate the effectiveness of our data synthesis pipeline, we synthesize data with OpenAI CUA\footnote{OpenAI CUA corresponds to the OpenAI \texttt{computer-use-preview} API 2025-03-01 release. We use Azure OpenAI for all API callling.} using queries from OpenWebVoyager~\citep{he2024openwebvoyager}. From these rollouts, we construct \webstar~(\textbf{Web}Voyager \textbf{S}tep-Level \textbf{T}rajectories with \textbf{A}ugmented \textbf{R}easoning), a dataset of 13.3K trajectories and 267K steps enriched with reasoning traces and step-level scores. To complement it, we further derive \webscore~(\textbf{Web}Voyager \textbf{S}tep-level \textbf{COR}rectness \textbf{E}valuation), a dedicated step-level grading dataset on 50K steps for training process reward models. We train Qwen-2.5-VL-Instruct-7B and 32B~\citep{Qwen2.5-VL} on \webstar~to assess the pipeline’s effectiveness. Our contributions are summarized as follows:
\begin{enumerate}
    \item \textbf{A scalable data synthesis pipeline with step-level filtering}: We propose a three-stage pipeline of trajectory collection, thought augmentation, and step-level filtering, which converts noisy rollouts into clean, reasoning-rich supervision. A key insight is that step-level filtering is essential for effective offline training, consistently outperforming trajectory-level filtering even under simple SFT. \looseness=-1
    \item \textbf{Empirical gains}: On live web tasks such as WebVoyager, our 7B model trained with step-level filtering surpasses the SoTA general-purpose CUA of similar size (\texttt{UI-TARS-1.5-7B}) by more than 15\%.
    \item \textbf{WebSTAR \& WebSCORE datasets}: We construct \webstar~for trajectory training with reasoning traces, and \webscore~for process reward modeling, providing high-quality resources for future CUA research.
    \item \textbf{StepRM}: We further train a lightweight 7B process reward model on \webscore, achieving grading quality comparable to \texttt{o4-mini} at a fraction of the cost, enabling scalable step-level filtering and paving way for RL-based optimization.
\end{enumerate}

\section{Preliminaries}
A CUA operates by receiving a task instruction and interacting with a graphical user interface to complete the task. At each step $i$, the agent observes the current state of the interface as a screenshot $o_i$, and executes an action $a_i$ in response. Given apriori knowledge of task difficulty, a maximum number of steps $T$ is defined, within which the agent must complete the task. To enhance the agent's reasoning and planning capabilities, it is common to adopt the ReAct framework~\citep{yaoreact}, which augments each step with an intermediate thought $t_i$ to guide the action $a_i$. To ensure broad generalizability, we do not provide the agent with auxiliary inputs such as accessibility trees or specialized annotations like Set-of-Marks used in prior work~\citep{he2024webvoyager,pahuja2025explorer}. 

The agent is trained to generate the next thought and action $(t_n,a_n)$, conditioned on the task instruction $\cI$, the sequence of past thoughts and actions $\{(t_{n-i},a_{n-i})\}_{i=1}^{n-1}$, and a truncated window of the most recent $w$ screenshots $\{o_{n-i}\}_{i=1}^w$. Only the image inputs are truncated due to their high token cost, which often dominates the input length in multimodal models. The action space is specified \Cref{tab:action_space}, where we generally follow common desktop operations defined in UI-TARS~\citep{qin2025ui}. \looseness=-1

During supervised finetuning (SFT), the training objective is to predict the correct thought and action in each step. Formally, given a dataset of trajectories: $\mathcal{D}=\{(\cI, \{o_i,t_i,a_i\}_{i=1}^T)\}$, the SFT objective minimizes the negative log-likelihood of the ground-truth thought–action pairs:\looseness=-1
\begin{align} \label{eq:sft_loss}
 \mathcal{L}_{\text{SFT}}(\theta) = 
 - \mathbbm{E}_{\mathcal{D}} \Bigg[
\sum_{n=1}^T
\log \pi_\theta(t_n, a_n \mid \cI,  \nonumber \\
\{t_{n-i}, a_{n-i}\}_{i=1}^{n-1}, \{o_{n-i}\}_{i=1}^w)
\Bigg].
\end{align}
This formulation makes clear that \textit{training operates at the step level}, aligning the supervision objective with the recursive nature of CUA execution. Unlike trajectory-level learning, which treats entire rollouts as atomic units, this step-level objective allows the agent to focus on producing locally correct behaviors at each decision point.

\begin{table}[t!]
\centering
\caption{Action space for computer use agents.}
\scalebox{0.7}{
\begin{tabular}{ll}
\toprule
\textbf{Action} & \textbf{Definition} \\
\midrule
\texttt{click(x,y)} & Clicks at coordinates $(x,y)$ \\
\texttt{left\_double(x,y)} & Double-clicks at $(x,y)$ \\
\texttt{right\_single(x,y)} & Right-clicks at $(x,y)$ \\
\texttt{drag(x1,y1,x2,y2)} & Drags from $(x_1,y_1)$ to $(x_2,y_2)$ \\
\texttt{scroll(x,y,dir)} & Scrolls at $(x,y)$ in given direction \\
\texttt{type(content)} & Types text \\
\texttt{hotkey(keys)} & Presses hotkey \\
\texttt{wait()} & Pauses 5s \\
\texttt{finished(content)} & Complete task with final answer \\
\bottomrule
\end{tabular}}
\vspace{-0.3cm}
\label{tab:action_space}
\end{table}

\section{Method}

\subsection{Teacher CUA Rollout}
We use queries from the training split of OpenWebVoyager~\citep{he2024openwebvoyager} and collect trajectories by rolling out the OpenAI \texttt{computer-use-preview} API. For each query, we generate 16 rollouts with a maximum horizon of 100 steps, resulting in 21K raw trajectories across 1,326 queries. Following prior works~\citep{he2024webvoyager,xue2025illusion}, we perform first-round trajectory-level filtering to remove failed rollouts by providing GPT-4o~\citep{hurst2024gpt} with the full action sequence and associated screenshots and prompting it to judge task success. After this initial grading, 13,338 successful trajectories are retained.


\subsection{Step-Level Filtering} \label{sec:step_filtering}

Previous attempts at data synthesis for CUAs often only retain the trajectories that successfully complete the task~\citep{murty2024bagel,pahuja2025explorer}. However, the trajectory-level approach does not align well with the \textit{step-level} nature of CUA training, as in Eq. \ref{eq:sft_loss}. The training objective focuses on predicting the next thought and action at each individual step, rather than modeling the entire trajectory end-to-end. The key limitation of trajectory-level filtering is that even successful trajectories often contain suboptimal or incorrect intermediate steps. Directly using these full trajectories for SFT exposes the model to flawed decision points, which can bias it toward replicating incorrect behaviors and ultimately degrade downstream performance. \looseness=-1


\begin{table*}[t!]
\caption{\textbf{Comparison of trajectory-level vs. step-level filtering in SFT.} Step-level filtering trains only on correct steps within successful trajectories, resulting in about half the data volume but consistently higher performance.}
\centering
\scalebox{0.65}{
\begin{tabular}{llrrrrrrrrrrrr}
\toprule
Trajectory & Step    & \# data & Allrecipes & Amazon & Apple & Arxiv & BBCNews & Coursera & ESPN & GitHub & Huggingface & Wolfram & Average \\
\midrule
Correct    & All     & 97K   & 48.9       & 26.8   & 32.6  & 16.3  & 28.6    & 45.2     & 25.0 & 19.5   & 19.0        & 37.0    & 29.9    \\
Correct    & Correct & 46K   & \textbf{60.0}       & \textbf{34.1}   & \textbf{41.9}  & \textbf{20.9}  & \textbf{47.6}    & \textbf{61.9}     & \textbf{29.5} & \textbf{31.7}   & \textbf{20.9}        & \textbf{47.8}    & \textbf{39.6}   \\
\bottomrule
\end{tabular}}
\label{tab:correct_step}
\end{table*}

\paragraph{Grading} To address the limitations of trajectory-level filtering, we propose a step-level filtering strategy that further evaluates each individual action in the successful trajectories. Instead of retaining entire trajectories, our method assigns a quality score to each step based on its expected contribution to task success. This evaluation is performed by a grading model that reasons over the context including the task instruction, the current and past actions, a window of the past $w$ screenshots annotated with actions following \citet{chen2024guicourse}, and a zoomed-in crop of the current screenshot centered on the action target, illustrated in \Cref{fig:main}. We use \texttt{o4-mini} as our grading model due to its strong reasoning capability and relatively lightweight inference cost. Importantly, \texttt{o4-mini} provides structured intermediate analysis, which enables more consistent judgments across diverse task types. The grading process proceeds in four stages: \looseness=-1
\begin{enumerate}
    \item \textbf{Screenshot analysis:} The grading model receives the past $w$ screenshots along with action annotations, and a zoomed-in crop of the latest screen focused on the relevant web element. This provides contextual grounding for understanding the task state.
    \item \textbf{Proposed action review:} The model rephrases the candidate action in natural language and judges whether it meaningfully contributes to the goal. Misaligned or unhelpful actions (e.g., misclicks) receive lower scores.
    \item \textbf{Alternative analysis:} The model proposes three alternative actions and simulates their likely outcomes. If any alternative action is strictly more effective\footnote{For instance, if the answer is already in view, reporting the answer directly is a strictly better action than scrolling.} than the proposed one, the proposed action is penalized.
    \item \textbf{Evaluation:} Based on correctness (from step 2) and optimality (from step 3), the model assigns an integer score from 0 to 10. A score of 0 indicates an irreversible error or guaranteed failure; 5 denotes a partially correct or suboptimal action; and 10 reflects an unambiguously helpful step with no superior alternative.
\end{enumerate}
An example structure of the response is provided in the step-level grading box in \Cref{fig:main}, and the detailed prompts describing grading criteria is presented in \Cref{appendix:prompt}. 

While \texttt{o4-mini} offers a practical balance between accuracy and efficiency, it remains a proprietary model. To make grading scalable, we construct \webscore, a dedicated step-level dataset with 200K steps where the inputs mirror those provided to \texttt{o4-mini}, and the labels are its outputs. In \Cref{sec:steprm}, we introduce an efficient alternative, \steprm, a 7B model trained on \webscore, which achieves grading accuracy on par with \texttt{o4-mini} while being significantly cheaper to deploy, making it suitable for large-scale data synthesis. \looseness=-1

\paragraph{Filtering} To integrate step-level grading into SFT, we retain all steps in the trajectory as contextual information but restrict the loss computation to only high-quality steps. Let $s_n \in \{0,1,\dots,10\}$ denote the score assigned by the grading model to step $n$. We define a binary mask $m_n=\mathbbm{1}\{s_n > 5\}$, where $m_n=1$ indicates that the step is judged correct and $m_n=0$ otherwise. The training input to the policy $\pi_\theta$ remains the full sequence of past thoughts, actions, and screenshots, ensuring that the model is exposed to realistic contexts that may include suboptimal decisions. However, \textit{the SFT objective only applies to the correct steps}:
\begin{align*}
\mathcal{L}_{\text{SFT}}(\theta) 
=& - \sum_{n=1}^T m_n \,
   \log \pi_\theta \Big(
      t_n, a_n \,\Big|\,
      \cI,  \\
      &\{t_{n-i},a_{n-i}\}_{i=1}^{n-1}, \{o_{n-i}\}_{i=1}^w
   \Big).
\end{align*}
This formulation ensures that incorrect actions remain part of the conditioning context, allowing the agent to learn recovery strategies, while preventing them from directly contributing to the optimization signal. In practice, we find that this selective masking greatly improves training stability and downstream task performance compared to trajectory-level filtering.

\paragraph{Less is more for CUA SFT}~Surprisingly, in the set of successful trajectories we collect, \textit{fewer than half of the steps are actually correct}. To quantify the impact of removing incorrect intermediate steps, \Cref{tab:correct_step} compares trajectory-level filtering (training on all steps from correct trajectories) with step-level filtering (training only on the correct steps within those trajectories). Although the step-level filtered dataset is roughly half the size of the trajectory-level one, it consistently yields substantial performance gains across all tasks. This result underscores that removing noisy steps is more beneficial than increasing dataset size for SFT.

\subsection{Thought Augmentation} \label{sec:thought}
The raw trajectories collected from OpenAI CUA consist solely of sequences of observations and actions, \textit{without any explicit reasoning traces}. However, prior work has shown that intermediate reasoning steps are crucial for enhancing agents' planning and decision-making capabilities, especially in complex, multi-step tasks~\citep{yaoreact}.

To address this, we follow previous approaches~\citep{xuaguvis,yang2024react,qin2025ui} and generate thoughts in a post-hoc manner. Each thought is designed to bridge perception and action, and is structured around three key components: \textbf{i) Situation description:} A brief description of the current observation, especially focusing on the region where the action takes place. \textbf{ii) Reasoning:} An explanation of the rationale behind the action and how it contributes to completing the overall task. \textbf{iii) Actionable instruction:} A natural language command that describes the action in an interpretable way, serving as a textual bridge between thought and execution. The detailed prompt is presented in \Cref{appendix:prompt}.

\begin{table}[t!]
\centering
\caption{\textbf{Comparison of training with and without thought augmentation.} The metrics are Pass@1 (Pass@4). Thought augmentation yields notably higher Pass@4, suggesting more reliable performance. \looseness=-1}
\scalebox{0.7}{
\begin{tabular}{l lcc}
\toprule
\textbf{Model} & \textbf{Task}
& \begin{tabular}[c]{@{}c@{}}\textbf{Correct steps} \\ w/o thought\end{tabular}
& \begin{tabular}[c]{@{}c@{}}\textbf{Correct steps} \\ w/ thought\end{tabular} \\
\midrule
\multirow{3}{*}{7B} 
 & WebVoyager       & \textbf{44.8} (57.2) & 43.3 (\textbf{64.2}) \\
 & Mind2Web-Live    & \textbf{23.0} (28.4) & 17.0 (\textbf{38.6}) \\
 & Online-Mind2Web  & \textbf{22.8} (27.7) & 17.0 (\textbf{29.4}) \\
\midrule
\multirow{3}{*}{32B} 
 & WebVoyager       & \textbf{48.6} (60.6) & 45.7 (\textbf{65.8}) \\
 & Mind2Web-Live    & \textbf{24.8} (37.5) & 20.4 (\textbf{39.7}) \\
 & Online-Mind2Web  & 23.2 (31.3) & \textbf{23.8 (38.1)} \\
\bottomrule
\end{tabular}}
\label{tab:thought}
\end{table}

\paragraph{Thought augmentation enhances robustness.}
The results in Table~\ref{tab:thought} show that incorporating thought augmentation yields consistent benefits, especially under the Pass@4 metric. In line with prior findings~\citep{xuaguvis}, the Pass@1 scores with and without thought traces remain close, which is expected since Pass@1 is a brittle indicator of reliability: CUA policies are stochastic, and in long-horizon tasks, a single error can derail an entire trajectory. By contrast, Pass@4 provides a more faithful measure of performance, reflecting whether the agent can succeed across multiple sampled rollouts rather than relying on a single attempt. \looseness=-1

Across benchmarks and model sizes, Pass@4 consistently improves with thought augmentation, indicating that reasoning traces substantially improve robustness. By explicitly linking perception to action through situation descriptions, rationales, and actionable instructions, thought augmentation helps reduce variance in decision-making. This results in more stable execution, more consistent task completion, and greater reliability when deploying CUAs in real-world environments.



\begin{figure*}[t!]
    \centering
    \begin{minipage}{0.65\linewidth}
        \begin{subfigure}[t]{0.5\textwidth}
        \centering
        \includegraphics[width=\linewidth]{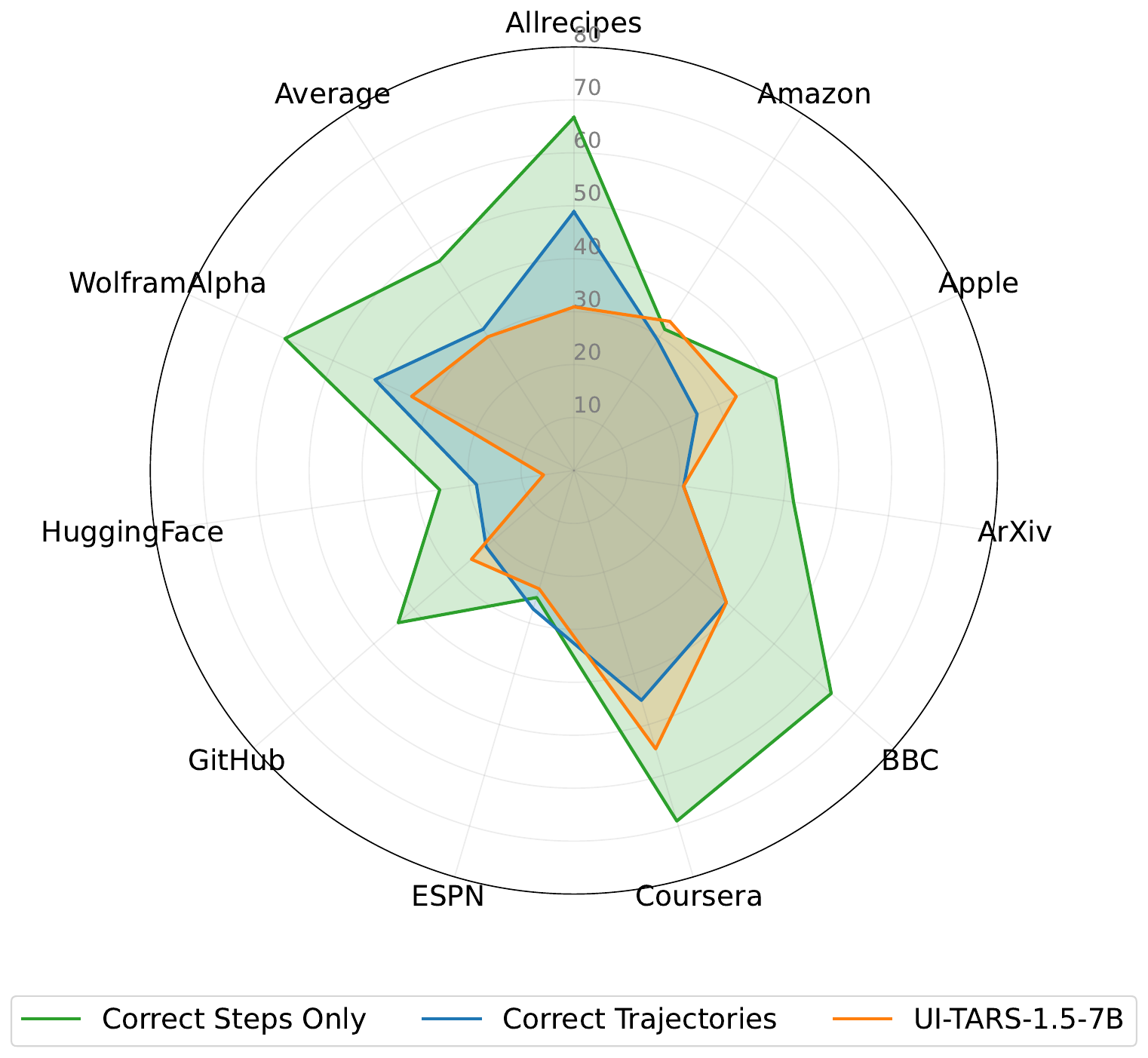}
    \end{subfigure}%
    ~
    \begin{subfigure}[t]{0.45\textwidth}
        \centering
        \includegraphics[width=\linewidth]{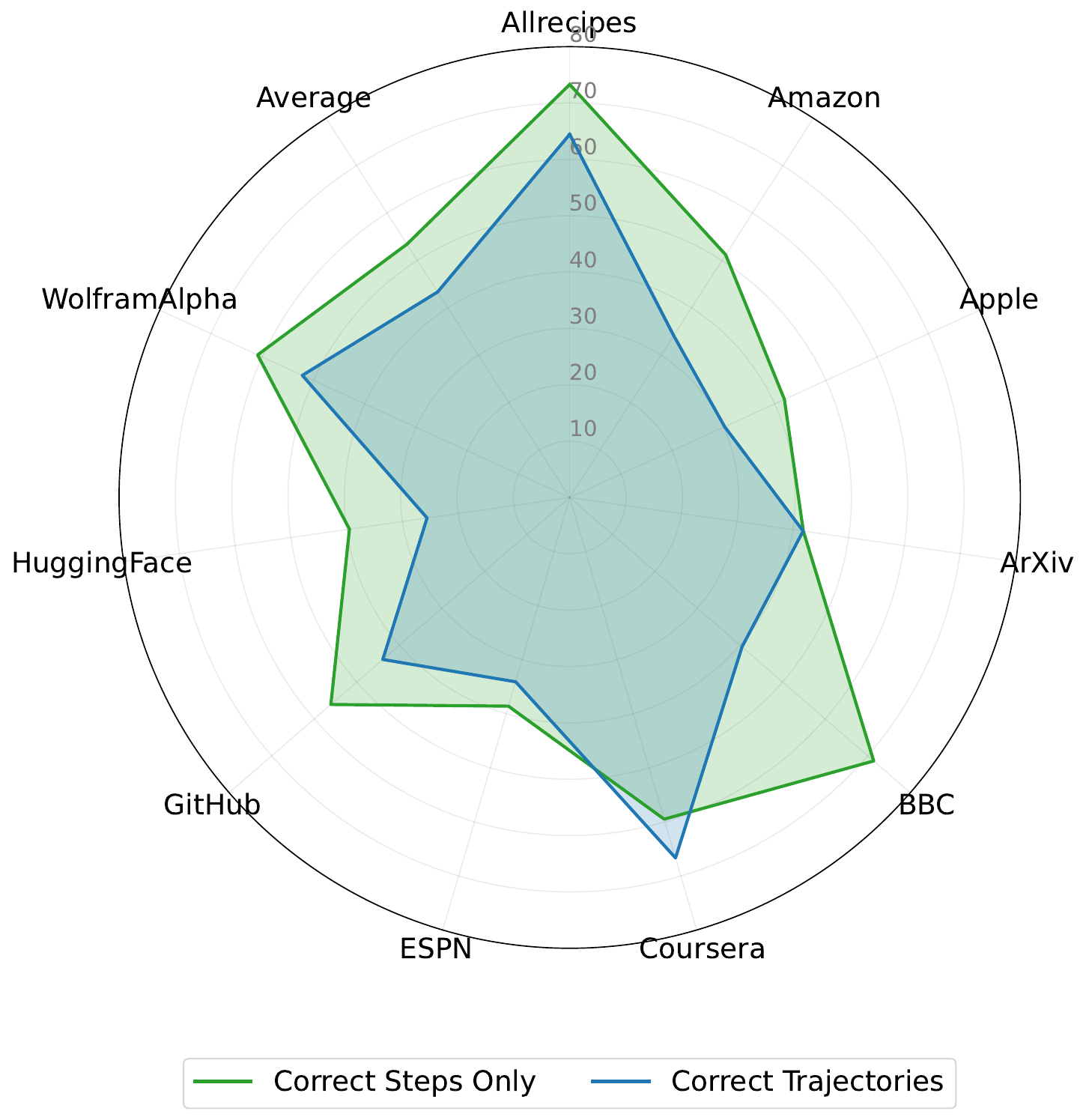}
    \end{subfigure}
    \caption{\textbf{WebVoyager results for different configurations.} Step-level filtering consistently outperforms trajectory-level filtering and even surpasses the SoTA general-purpose CUA \texttt{UI-TARS-1.5-7B}, despite using only SFT.}
    \label{fig:webvoyager}
    \vspace{-0.3cm}
    \end{minipage}
    \begin{minipage}{0.33\textwidth}
        \centering
        \vspace{0pt} 
        \captionof{table}{\textbf{Performance on Mind2Web suites with Pass@1 (Pass@4).} Step-level filtering substantially improves success rates across both Mind2Web-Live and Online-Mind2Web, with larger gains for 32B models. This confirms that cleaner step-level supervision benefits models of all scales.}
        \scalebox{0.5}{\begin{tabular}{cccc}
\toprule
\textbf{Model Size} & \textbf{Tasks} & \textbf{All steps} & \textbf{Correct steps} \\
\midrule
\multirow{2}{*}{7B} 
 & Mind2Web-Live    & 15.0 (25.0) & \textbf{17.0 (38.6)} \\
 & Online-Mind2Web  & 12.0 (23.0) & \textbf{17.0 (29.4)} \\
\midrule
\multirow{2}{*}{32B} 
 & Mind2Web-Live    & 14.5 (26.1) & \textbf{20.4 (39.7)} \\
 & Online-Mind2Web  & 14.5 (24.8) & \textbf{23.8 (38.1)} \\
\bottomrule
\end{tabular}\label{tab:mind2web}}
    \end{minipage}
\end{figure*}

\section{Experiments}

\subsection{Setup}

\paragraph{Data synthesis}~We roll out OpenAI CUA on queries from the OpenWebVoyager~\citep{he2024openwebvoyager} dataset. The original query set comes from the training split of both WebVoyager~\citep{he2024webvoyager} and Mind2Web~\citep{deng2023mind2web}, totaling 2,111 queries. However, due to the dynamic nature of websites, many tasks become outdated or invalid, caused by issues such as websites access errors, added human verification steps, or updated URLs. After filtering out these queries, we retain 1,326 valid ones. Details about those website failures are discussed in \Cref{appendix:eval}.

To interact with live websites, we adopt the WebVoyager~\citep{he2024webvoyager} codebase. However, the original implementation only supports Set-of-Mark (SoM) prompting, which is incompatible with the more common raw coordinate interaction. Thus, we modify the implementation with Playwright to enable more robust interactions with websites. For each query, we cap the rollout length at 100 steps (i.e., $T=100$), terminating the episode at step 100 regardless of whether the task is completed. \looseness=-1

\paragraph{Training}~We use \texttt{Qwen-2.5-VL-7B-Instruct} and \texttt{Qwen-2.5-VL-32B-Instruct}~\citep{Qwen2.5-VL} as base models. Following common practice~\citep{wang2024qwen2,xuaguvis}, we freeze the vision tower and only train the language components. All experiments are conducted on NVIDIA A100 GPUs with 80 GB memory, using LlamaFactory~\citep{zheng2024llamafactory} for multimodal SFT. We set the sequence length to 8,192 and feed in only the most recent screenshot ($w=1$) following \citet{xuaguvis}, as this configuration yields the most stable training in our setting. We use an initial learning rate of 1e-5 with warmup ratio of 0.03 and cosine decay to 0. The global batch size is set to 128. Training on \webstar~for 2 epochs takes approximately 256 GPU hours for 7B models and 1024 GPU hours for 32B models. \looseness=-1

\paragraph{Evaluation}~To test the capabilities of CUA, we deploy three online evaluation benchmarks: WebVoyager~\citep{he2024webvoyager}, Mind2Web-Live~\citep{pan2024webcanvas} and Online Mind2Web~\citep{xue2025illusion}. The online evaluation is particularly challenging due to the everchanging nature of websites, requiring the model to generalize the static data during offline training to dynamically changing environments. Following \citet{he2024webvoyager}, we provide the full action trajectory and corresponding screenshots of the agent to GPT-4o~\citep{hurst2024gpt} to automatically judge the task success.

\subsection{Main Results}

The results in \Cref{fig:webvoyager} and \Cref{tab:mind2web} highlight the clear advantages of step-level filtering for CUA SFT. In the WebVoyager benchmark, training on only the correct steps in correct trajectories consistently yields stronger performance across almost all domains compared to training on all steps within correct trajectories. Under this setting, our 7B step-level model achieves an average success rate of 47.0\%, substantially outperforming the SoTA general-purpose CUA \texttt{UI-TARS-1.5-7B} (30.0\%) despite using only SFT. This demonstrates that carefully filtered training data can rival and even surpass the performance of stronger baselines trained with more complex methods. The full numerical results and statistical significance test are presented in \Cref{appendix:results}.

The same pattern holds on the Mind2Web evaluations. Across both Mind2Web-Live and Online-Mind2Web, step-level filtering substantially improves success rates for both 7B and 32B models. The gains are larger for the 32B models, indicating that larger models can capitalize more effectively on clean supervision once noisy steps are removed, further amplifying the benefits of filtering.

Together, these findings demonstrate that step-level filtering is critical for effective offline training of CUAs. By removing incorrect or suboptimal intermediate actions while preserving correct steps from successful trajectories, the resulting training data better aligns with the step-level prediction objective. This leads to significant improvements in downstream performance, offering a simple yet powerful strategy for advancing the robustness and efficiency of CUA training.

\begin{figure}[t!]
    \centering
    \includegraphics[width=0.8\linewidth]{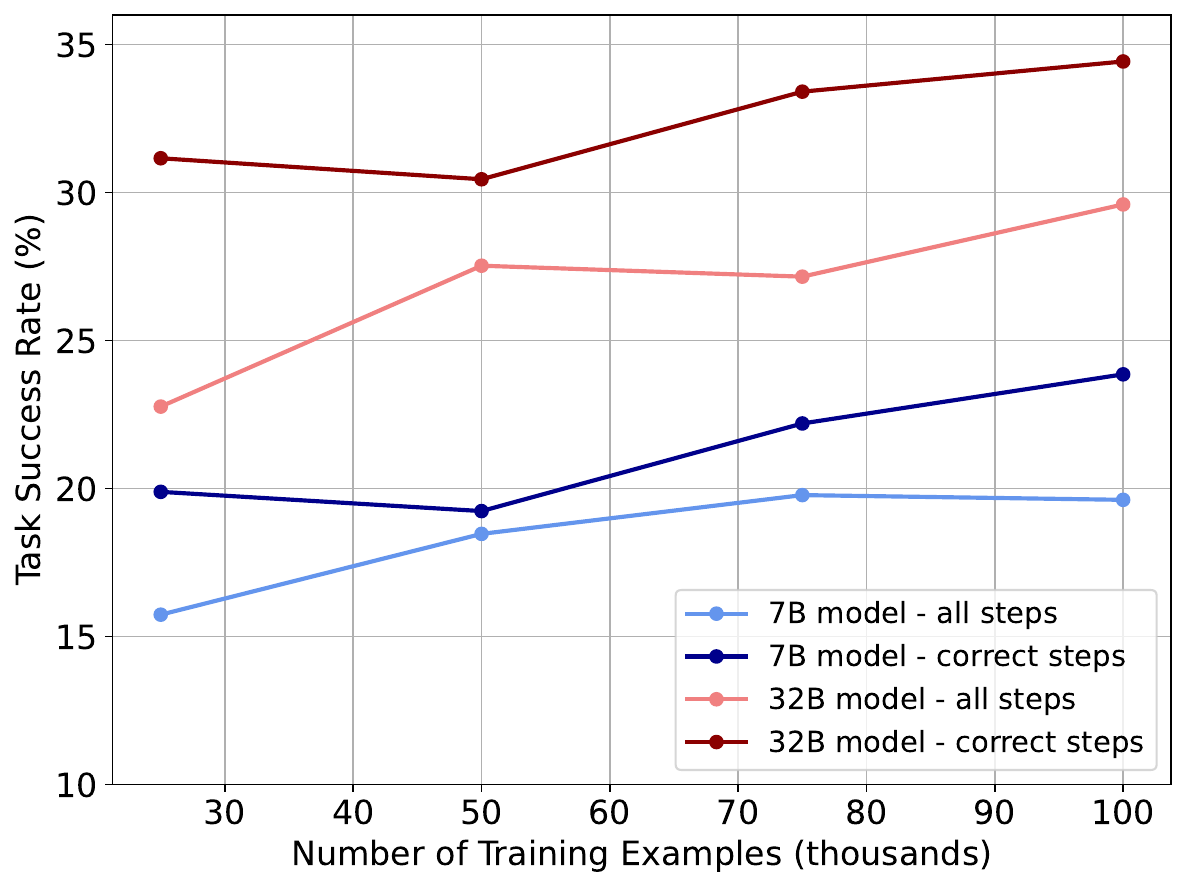}
    \caption{\textbf{Scaling behavior of step-level vs. trajectory-level filtering.} Step-level filtering consistently outperforms trajectory-level filtering for both 7B and 32B models, with clear gains in low-data settings and stronger improvements as dataset size increases.}
    \label{fig:data_scale}
    \vspace{-0.3cm}
\end{figure}

\paragraph{Performance with data scaling}~\cref{fig:data_scale} shows the trend of average performance across the three benchmarks with increasing amount of data under different model sizes. The results highlight that step-level filtering not only improves performance but also scales more effectively with additional data compared to trajectory-level filtering. In the low-data regime (around 25K examples), models trained on correct steps already achieve substantially higher success rates than those trained on full trajectories. As the number of examples increases, the advantage of step-level filtering becomes even more pronounced: both the 7B and 32B models continue to gain steadily, while trajectory-level filtering plateaus early, particularly for the 7B model.

The scalability benefits are more pronounced for the 32B model, showing that larger models can make even better use of clean supervision with more data. By contrast, the trajectory-level baseline falls short in capitalizing on additional data, highlighting the inefficiency of learning from noisy trajectories. Overall, these results show that step-level filtering provides a dual advantage: it yields strong gains when data is scarce and continues to deliver improvements as the dataset grows, making it a robust and scalable strategy for training CUAs. \looseness=-1

\subsection{Ablations} \label{sec:ablations}
In this section, we further validate the consistency and reliability of \texttt{o4-mini} grading.

\begin{table}[t!]
\centering
\caption{Confusion matrix for grading.} \label{tab:conf_mat}
\scalebox{0.6}{
\begin{tabular}{lcc}
\toprule
 & \texttt{o4-mini} $\geq 5$ & \texttt{o4-mini} $< 5$ \\
\midrule
Human $\geq 5$ & 36 & 16 \\
Human  $< 5$ & 11 & 37 \\ 
\bottomrule
\end{tabular}}
\end{table}

\paragraph{Human score validations} To verify that \texttt{o4-mini’s} step-level scores reflect meaningful correctness signals, we manually annotated 100 randomly sampled steps and compared them against the model’s judgments. Using a binary agreement criterion (both assigning scores $\geq 5$ or $<5$), we observed a $73\%$ agreement rate. We find that the main source of disagreement occurs when the model penalizes exploratory-but-not-wrong steps more harshly than a human annotator would. This behavior is expected: the grading prompt is deliberately strict to prefer clearly goal-directed actions, and we refined this prompt through extensive iterations to achieve the most consistent downstream performance. The confusion matrix (\Cref{tab:conf_mat}) also shows that most disagreements are false negatives. Overall, the agreement analysis supports that the step-level scores are aligned with human intuition and do not exhibit systematic bias that would undermine the filtering process.

\paragraph{Grading consistency} To assess the stability of the grader, we randomly sampled ~230 steps from our training data and queried \texttt{o4-mini} five times per step. We then computed score variance across the repeated samples. On average, scores varied by only 0.81 points, with a median standard deviation of 0.49. The coefficient of variation was similarly low (median $<6\%$), indicating that \texttt{o4-mini}’s outputs are generally stable and consistent across repeated evaluations.

\begin{table}[t!]
\centering
\caption{Ablation on score cutoff threshold on 7B model with metrics Pass@1 (Pass@4).} \label{tab:cutoff}
\scalebox{0.6}{
\begin{tabular}{lccccc}
\toprule
Cutoff & 2 & 4 & 5 & 6 & 8 \\
\midrule
WebVoyager &  29.7 (53.7) &	40.2 (64.9)&	43.3 (64.2)&	41.3 (64.2)&	42.5 (62.7) \\
\bottomrule
\end{tabular}}
\end{table}

\paragraph{Score threshold} In our experiment, we choose 5 as the step-grading cutoff, as a score of 5 is explicitly defined as the boundary case for a partially correct step. To further justify this choice, we conducted an ablation over grading thresholds \{2, 4, 6, 8\}. For each threshold, we performed filtering and kept the same final dataset size (100K steps) for a fair comparison. The full results are shown in the table below; we report Pass@1 and Pass@4 in parentheses. We can see that threshold = 5 achieves competitive performance. Note that using higher thresholds results in lower initial retention rates, requiring many more rollouts to reach 100K steps, making them significantly less scalable. Overall, these results indicate that threshold = 5 strikes a practical balance between data quality (removing noisy steps) and data efficiency (reasonable retention). \looseness=-1

\subsection{StepRM} \label{sec:steprm}

As discussed in \Cref{sec:step_filtering}, step-level filtering is central to our pipeline: each action in a trajectory is graded individually to decide whether it provides useful supervision. In our primary setup, we relied on \texttt{o4-mini} as the grading model, given its strong reasoning ability and structured outputs. However, \texttt{o4-mini} is proprietary and relatively expensive to query, especially since step-level grading requires evaluating up to $T=100$ actions per trajectory, each with screenshot context and action history.

To alleviate this bottleneck, we introduce \steprm, a lightweight multimodal process reward model based on \texttt{Qwen2.5-VL-7B}. \steprm~is explicitly trained to replicate the step-level grading behavior of \texttt{o4-mini}, effectively distilling its judgments into a compact, open model. The training is conducted on the \webscore~dataset, which contains 200K graded steps randomly sampled from the raw rollout trajectories collected in our data synthesis pipeline. To ensure balanced supervision, the dataset contains 50\% steps with score $>5$ and 50\% with score $\leq 5$. Each sample includes the same multimodal inputs used in our grading pipeline (the latest annotated screenshot, recent history of actions, and task instruction) and predicts a scalar quality score for the current action. In this way, \steprm~functions as a direct substitute for \texttt{o4-mini} in step-level filtering, while being significantly cheaper to deploy at scale. \steprm uses the same hyperparameters and compute setup as our main policy models. \looseness=-1

\begin{table}[t!]
\centering
\caption{\textbf{Step-level filtering performance from StepRM with Pass@1 (Pass@4).} Compared with o4-mini as a grading model, \steprm~achieves competitive quality while being substantially more efficient, demonstrating that lightweight learned process reward models can replace expensive LLM evaluations for step-level filtering.}
\scalebox{0.6}{
\begin{tabular}{ccccc}
\toprule
 \textbf{Model Size} & \textbf{Tasks} & \textbf{All steps} & \begin{tabular}[c]{@{}c@{}}\textbf{Correct steps} \\ o4-mini\end{tabular} & \begin{tabular}[c]{@{}c@{}}\textbf{Correct steps} \\ \steprm \end{tabular} \\
 \midrule
&WebVoyager     & 27.0 (49.1) & \textbf{38.4 (63.7)} & 35.4 (57.9) \\
7B &Mind2Web-Live   & 6.3 (19.4)  & \textbf{18.5 (36.0)} & 13.2 (28.4) \\
&Online-Mind2web & 9.5 (20.3)  & \textbf{17.3 (31.2)} & 13.0 (24.4) \\
\midrule 
&WebVoyager     & 33.4 (58.3) & 45.4 (66.3) & \textbf{45.7 (67.4)} \\
32B &Mind2Weblive   & 8.9 (25.0)  & \textbf{20.5} (36.4) & 20.0 \textbf{(38.6)} \\
&OnlineMind2web & 13.1 (27.3) & \textbf{16.5 (29.2)} & 15.3 (29.1) \\
\bottomrule
\end{tabular}}
\label{tab:steprm}
\end{table}

For testing the performance of StepRM, we take a separate subset of 50K steps. As shown in Table~\ref{tab:steprm}, both \texttt{o4-mini} and \steprm~filtering yield large improvements over trajectory-level filtering, confirming that removing noisy steps is crucial for CUA training. For a fair comparison, both graders start from an independently collected 200K-step dataset distinct from the \webscore~training set and perform filtering on this shared pool before SFT. Under this controlled setup, \steprm~achieves performance on par with \texttt{o4-mini} across multiple benchmarks, while being far more efficient to deploy at scale. \looseness=-1

Beyond serving as a cost-effective substitute for step-level filtering, \steprm~also has the potential to act as a critic in reinforcement learning frameworks such as PPO~\citep{schulman2017proximal}. In this setting, \steprm~can provide dense, step-level feedback to the agent, helping to overcome two central challenges in CUA training: sparse rewards, where task success is only known at the end of long trajectories, and credit assignment, where it is difficult to determine which intermediate actions led to success or failure. By offering fine-grained step evaluations, \steprm~could enable more sample-efficient RL training of CUAs in interactive environments. We leave the integration of \steprm~into RL pipelines as an exciting direction for future work. \looseness=-1


\section{Related Works}

\paragraph{Computer use agent datasets}~Computer use agents (CUAs) have recently emerged as a powerful class of autonomous systems that can operate directly on graphical user interfaces~\citep{operator,Claude-computer-use,agasheagent,qin2025ui,wang2025ui,wang2025opencua}. Training these agents is particularly challenging because it requires grounding visual input into actionable affordances, reasoning and planning over long horizons, and executing precise actions in dynamic environments. While direct online interaction with GUIs would be the most natural way to acquire these capabilities, it is prohibitively expensive due to the need for virtual machines and complex infrastructure. As a result, most prior work relies on offline training with pre-collected trajectories.

Human-annotated datasets provide valuable supervision but are expensive and limited in scale. UI-TARS~\citep{qin2025ui} collected human demonstrations along with curated self-reflection examples, and UI-TARS-2~\citep{wang2025ui} went further by requiring annotators to verbalize their reasoning traces during trajectory collection. Others, including MM-Mind2Web~\citep{zheng2024gpt} and GUIAct~\citep{chen2024guicourse}, provide only short action trajectories without reasoning traces or step-level grading, while AgentTrek~\citep{xuagenttrek} includes reasoning but lacks explicit step quality control. Earlier datasets~\citep{liu2018reinforcement,yao2022webshop,koh2024visualwebarena} rely on Set-of-Marks or accessibility trees, which simplifies grounding but deviates from realistic deployment settings where only screenshots are available. These limitations motivate the need for scalable, reasoning-rich, and coordinate-based CUA data.

\begin{table}[t!]
\centering
\caption{Comparison with existing datasets.} \label{tab:comparison}
\scalebox{0.65}{
\begin{tabular}{lcccc}
\toprule
\textbf{Datasets} & \textbf{Avg. Steps} & \textbf{\# Traj} &  \textbf{Thought} & \textbf{Step Grading} \\
\midrule
MM-Mind2Web & 7.7 & 1K & \xmark & \xmark \\
GUIAct & 6.7 & 2.5K & \xmark & \xmark \\ 
AgentTrek & 12.1 & 10.4K & \cmark & \xmark \\
\midrule
WebSTAR (Ours) & 22.1& 13.3K & \cmark & \cmark \\
\bottomrule
\end{tabular}}
\end{table}

As summarized in Table~\ref{tab:comparison}, \webstar~is the first to combine scale, long-horizon trajectories, structured reasoning traces, and step-level grading, offering realistic and high-quality supervision without human annotation or auxiliary signals.

\paragraph{Data filtering}~With large-scale but noisy synthetic rollouts, filtering becomes crucial for ensuring that models learn from high-quality supervision. The dominant strategy in CUAs has been trajectory-level filtering, which retains only trajectories that reach successful task completion~\citep{sun-etal-2025-os}. While straightforward, this approach is mismatched to the step-wise nature of CUA training, as successful trajectories often include suboptimal or even incorrect intermediate steps.

Broader works in reinforcement learning have explored step-level filtering, most notably filtered behavior cloning (FBC)~\citep{gulcehre2023reinforced,singhbeyond,shao2024deepseekmath}, where the model selectively trains on high-quality actions. However, in CUAs, step-level filtering is especially challenging because rewards are inherently sparse: only final success is observed, and Monte Carlo credit assignment is high-variance. Actor-critic methods attempt to address this by learning a critic to provide intermediate feedback~\citep{bai2024digirl,bai2025digiq}, but these methods are often unstable and costly to train~\citep{parisi2019td}.

Our work advances this line by introducing a scalable step-level filtering strategy that relies on a separate grading model. Compared with actor-critic methods, our filtering is stable and cost-effective, while still aligning supervision with the step-level nature of CUA training. This makes our method both practically scalable and more faithful to the underlying training objective. \looseness=-1

\section{Conclusions}
In this work, we present a scalable data synthesis pipeline for training computer use agents, addressing the limitations of costly human annotation and noisy trajectory-level filtering. Our key insight is that step-level filtering better aligns with the supervision objective, enabling agents to learn from correct actions even within imperfect trajectories. Combined with thought augmentation, which strengthens planning and stability, this approach yields substantial improvements across multiple benchmarks, surpassing strong baselines with only SFT. To support broader research, we introduce two complementary datasets: \webstar, a trajectory dataset enriched with reasoning and step-level scores, and \webscore, a grading dataset tailored for training process reward models. Building on \webscore, we introduced \steprm, a lightweight process reward model that matches o4-mini in grading quality at a lower cost, offering a practical tool for future reinforcement learning based optimization. Together, these contributions establish step-level filtering, reasoning-rich augmentation, and efficient reward modeling as core principles for robust and scalable CUA training.

\newpage

\section*{Limitations}
While our work demonstrates that step-level filtering and thought augmentation substantially improve the scalability and reliability of computer-use agents (CUAs), several limitations remain.

Firstly, the scalability of our pipeline is partly constrained by the quality of the teacher CUA used for data collection. Even state-of-the-art CUAs introduce noise and suboptimal behaviors, which limits the diversity of high-quality steps that can be synthesized. As CUAs themselves improve, we expect this bottleneck to diminish, enabling higher-quality data generation at scale.

Secondly, our approach relies exclusively on supervised finetuning (SFT). Although effective, SFT alone cannot fully address issues such as long-horizon credit assignment or exploration in unseen environments. Reinforcement learning (RL) has the potential to improve these aspects, and our \steprm~critic model offers a natural pathway toward integrating step-level filtering with RL in future work.


\bibliography{custom}

\begin{table*}[t!]
\centering
\scalebox{0.85}{
\begin{tabular}{lccc|c}
\toprule
\textbf{Domain} & \textbf{All steps} & \textbf{Correct steps} & \textbf{UI-TARS-1.5-7B} & \textbf{OAI CUA} \\
\midrule
Allrecipes   & 48.9 (80.0) & \textbf{66.7 (86.7)} & 30.9 (56.8) & 78.2 (91.1) \\
Amazon       & 29.3 (46.3) & 31.7 (56.1) & \textbf{33.5 (65.9)} & 54.3 (92.7) \\
Apple        & 25.6 (46.5) & \textbf{41.9} (55.8) & 33.7 (\textbf{60.5}) & 68.6 (79.1) \\
Arxiv        & 20.9 (34.9) & \textbf{41.9 (58.1)} & 20.9 (39.5) & 71.5 (90.7) \\
BBCNews      & 38.1 (61.9) & \textbf{64.3} (73.8) & 38.1 (\textbf{76.2}) & 67.1 (81.0) \\
Coursera     & 45.2 (66.7) & \textbf{69.0 (78.6)} & 54.8 (76.2) & 79.8 (92.9) \\
ESPN         & \textbf{27.3 (47.7)} & 25.0 (50.0) & 23.3 (43.2) & 56.8 (75.0) \\
GitHub       & 21.9 (75.6) & \textbf{43.9 (82.9)} & 25.6 (48.8) & 66.5 (85.4) \\
Huggingface  & 18.6 (25.6) & \textbf{25.6 (34.9)} & 5.8 (11.6)  & 61.6 (79.1) \\
Wolfram      & 41.3 (\textbf{65.2}) & \textbf{60.0 (65.2)} & 33.7 (47.8) & 85.3 (93.5) \\
\midrule
\textbf{Average} & 31.9 (55.1) & \textbf{47.0 (64.2)} & 30.0 (52.4) & 69.2 (86.0) \\
\bottomrule
\end{tabular}}
\caption{\textbf{Performance comparison on 7B models}. The reported metrics are Pass@1 and Pass@4 (in parentheses). The bold numbers show the best performing metric among the 7B models, where OAI CUA serves as the oracle. Step-level filtering (“Correct steps”) substantially improves over trajectory-level filtering (“All steps”), and further outperforms \texttt{UI-TARS-1.5-7B} across domains. OpenAI CUA (OAI CUA) remains strongest, but our method narrows the gap significantly.}
\label{tab:comparison_7b}
\end{table*}

\newpage
\appendix

\section{More Results} \label{appendix:results}

\begin{table}[t!]
\centering
\scalebox{0.85}{
\begin{tabular}{lcc}
\toprule
\textbf{Domain} & \textbf{All steps} & \textbf{Correct steps} \\
\midrule
Allrecipes   & 64.4 (84.4) & \textbf{73.3 (91.1)} \\
Amazon       & 34.1 (41.5) & \textbf{48.8 (51.2)} \\
Apple        & 30.2 (44.2) & \textbf{41.9 (62.8)} \\
Arxiv        & 41.9 (44.2) & \textbf{41.9 (53.5)} \\
BBCNews      & 40.5 (61.9) & \textbf{71.4 (76.2)} \\
Coursera     & \textbf{66.7} (76.2) & 59.5 (\textbf{85.7}) \\
ESPN         & 34.1 (43.2) & \textbf{38.6 (50.0)} \\
GitHub       & 43.9 (73.2) & \textbf{56.1 (75.6)} \\
Huggingface  & 25.6 (30.2) & \textbf{39.5 (46.5)} \\
Wolfram      & 52.2 (\textbf{69.6}) & \textbf{60.9} (67.4) \\
\midrule
\textbf{Average} & 43.5 (56.9) & \textbf{53.5 (65.8)} \\
\bottomrule
\end{tabular}}
\caption{\textbf{Performance comparison for 32B models.} Step-level filtering (“Correct steps”) consistently improves over trajectory-level filtering (“All steps”), yielding large gains across nearly all domains.}
\label{tab:32b_results}
\end{table}

\paragraph{Full numerical results} We present the full numerical results of WebVoyager in \Cref{tab:comparison_7b} and \Cref{tab:32b_results}. To reduce the variance in the evaluation process, we further report the Pass@4 metric beyond Pass@1. Under both model sizes, step-level filtering significantly outperforms trajectory level filtering, validating the effectiveness of our filtering strategy. OAI CUA still obtains the strongest performance, but our 32B model results demonstrated in \Cref{tab:32b_results} already closes the gap significantly. With the effective scaling of both data (shown in \Cref{fig:data_scale} and model, we believe our data synthesis pipeline can yield a recipe that approach OAI CUA performance.

There are domains where smaller models fail to close the gap with the teacher CUA, wuch as HuggingFace, and our analysis shows this domain is more challenging than others. The queries here typically require deep technical knowledge, multi-hop navigation, and precise extraction of information from code-heavy documentation (e.g., implementation details of TRL modules or tokenizer parameters). The teacher model (OAI CUA) is far more sample-efficient in locating the correct documentation sections, whereas the student often spends many turns navigating intermediate pages. A key observation is that failure is dominated by running out of turns, not by incorrect reasoning. Across HuggingFace queries, the student exhausts the 100-step limit in 65\% of tasks, compared to 28\% for OAI CUA. 

\begin{table*}[t!]
\centering
\scalebox{0.85}{
\begin{tabular}{lccc|c}
\toprule
\textbf{Domain} & \textbf{All steps} & \textbf{Correct steps} & \textbf{UI-TARS-1.5-7B} & \textbf{OAI CUA} \\
\midrule
Allrecipes      
& 49.4 $\pm$ 9.9 & \textbf{59.4 $\pm$ 9.9} & 30.9 $\pm$ 10.0 & 78.2 $\pm$ 9.9 \\
Amazon          
& 25.0 $\pm$ 9.7 & \textbf{34.8 $\pm$ 11.6} & 33.5 $\pm$ 9.9 & 54.3 $\pm$ 7.7 \\
Apple           
& 26.7 $\pm$ 10.9 & \textbf{36.0 $\pm$ 11.7} & 33.7 $\pm$ 10.3 & 68.6 $\pm$ 12.5 \\
Arxiv           
& 24.4 $\pm$ 11.2 & \textbf{34.3 $\pm$ 11.5} & 20.9 $\pm$ 9.7 & 71.5 $\pm$ 9.6 \\
BBCNews         
& 29.8 $\pm$ 9.1 & \textbf{45.8 $\pm$ 10.6} & 38.1 $\pm$ 9.2 & 67.1 $\pm$ 11.6 \\
Coursera        
& 48.2 $\pm$ 12.6 & \textbf{64.9 $\pm$ 12.0} & 54.8 $\pm$ 11.0 & 79.8 $\pm$ 9.1 \\
ESPN            
& 19.9 $\pm$ 7.8 & \textbf{31.2 $\pm$ 10.9} & 23.3 $\pm$ 9.1 & 56.8 $\pm$ 11.8 \\
GitHub          
& 40.9 $\pm$ 10.5 & \textbf{48.8 $\pm$ 10.5} & 25.6 $\pm$ 9.3 & 66.5 $\pm$ 10.9 \\
Huggingface     
& 12.2 $\pm$ 7.5 & \textbf{20.3 $\pm$ 10.0} & 5.8  $\pm$ 5.5 & 61.6 $\pm$ 11.7 \\
Wolfram         
& 47.8 $\pm$ 12.1 & \textbf{57.1 $\pm$ 13.2} & 33.7 $\pm$ 12.0 & 85.3 $\pm$ 8.1 \\
\midrule
\textbf{WebVoyager}
& 32.6 $\pm$ 3.5 & \textbf{43.4 $\pm$ 3.8} & 30.0 $\pm$ 3.3 & 69.2 $\pm$ 3.4 \\
\textbf{Mind2Web-Live}
& 14.8 $\pm$ 6.2 & 17.2 $\pm$ 4.6 & \textbf{18.1 $\pm$ 6.0} & 50.3 $\pm$ 8.5 \\
\textbf{Online-Mind2Web}
& 12.3 $\pm$ 3.9 & \textbf{17.0 $\pm$ 5.7} & 14.6 $\pm$ 4.5 & 48.5 $\pm$ 6.3 \\
\bottomrule
\end{tabular}}
\caption{\textbf{Performance with 95\% confidence intervals on 7B models.} Values are Pass@1 in percentages.}
\label{tab:comparison_ci}
\end{table*}

\paragraph{Statistical significance} We include the 95\% confidence interval for our main results on the 7B models in \Cref{tab:comparison_ci}. To compute error bounds, we run each query 4 times. These intervals show that the improvements we report, such as the gains from step-level filtering and the performance differences among compared methods, remain statistically significant under reasonable uncertainty estimates. In addition, in the paper, we already report Pass@4, which is a more stable and reliable metric for long-horizon CUA tasks and further reduces variance compared to Pass@1.

\begin{figure}[t!]
    \centering
    \includegraphics[width=0.8\linewidth]{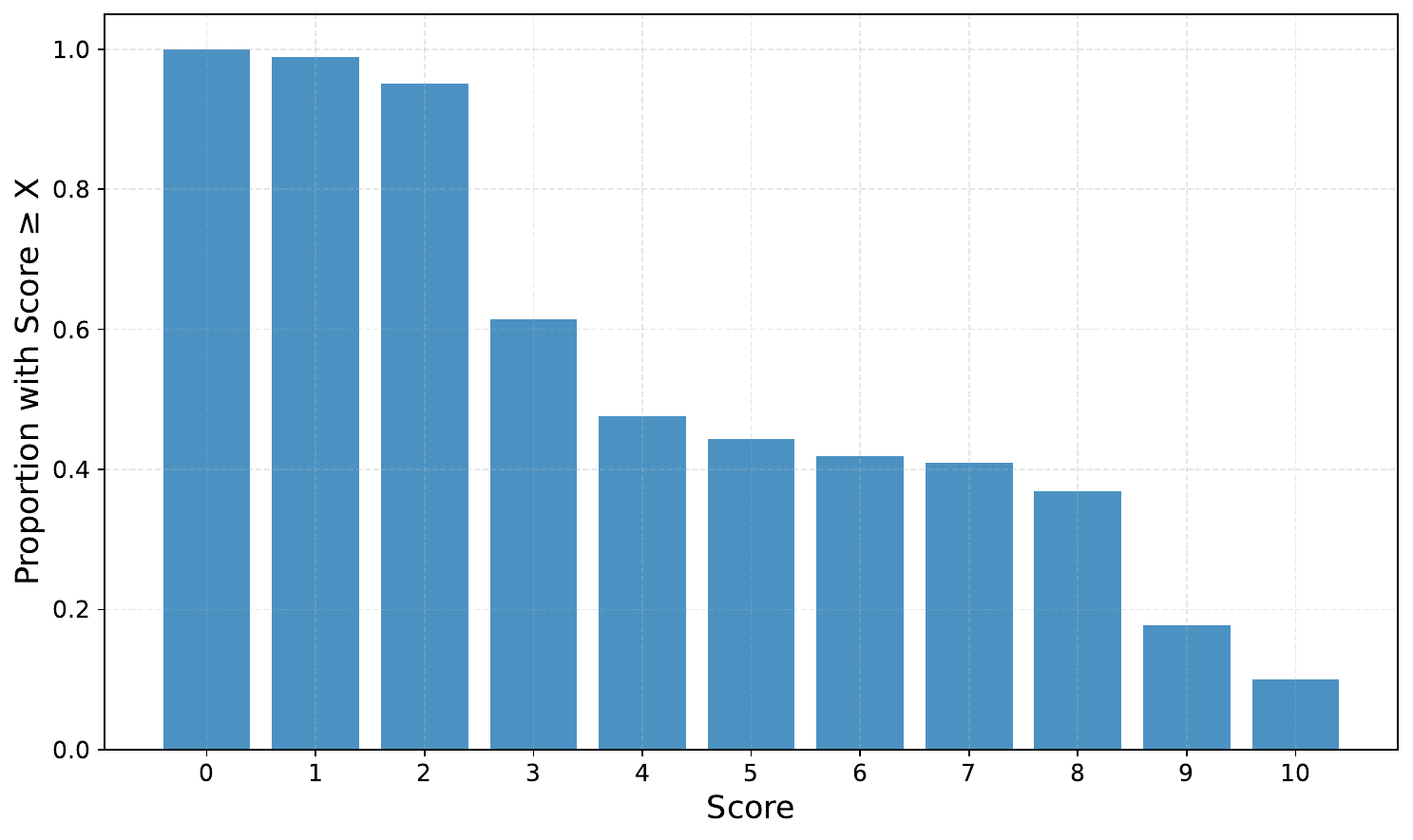}
    \caption{\textbf{Step-level score distribution from \texttt{o4-mini}.} The CDF shows the proportion of actions with scores above each threshold, highlighting that a large share of steps are graded as suboptimal. }
    \label{fig:score}
\end{figure}

\paragraph{Score distribution} The CDF in \Cref{fig:score} illustrates the cumulative distribution of step-level scores. The distribution is computed only over the successful trajectories, as our final goal is to only retain the correct steps in the successful trajectories. Even within those successful trajectories, a majority of actions receive low scores: over half fall below 5. This confirms that trajectory-level filtering, which accepts whole rollouts as long as the final task succeeds, inevitably includes many flawed intermediate steps. Such noisy supervision can mislead agents during training. Step-level filtering directly addresses this issue by isolating the small fraction of high-quality actions (scores close to 10) while discarding the majority of low-quality ones. This reinforces our key insight that filtering at the step granularity is essential for constructing cleaner and more reliable training data for CUAs.

\section{Evaluation Details} \label{appendix:eval}

\begin{table*}[t!]
    \centering
    \caption{Websites with access problems in Mind2Web queries.}
    \scalebox{0.9}{
    \begin{tabular}{p{3cm}p{10cm}}
    \toprule
    Error     &  Websites \\
    \midrule
    Human verification & \texttt{bbb.org, redfin.com, extraspace.com, seatgeek.com, google.com/shopping, sec.gov, expedia.com, discogs.com} \\
    Access denied & \texttt{americashealthrankings.org, adoptapet.com, reddit.com, qatarairways.com, fedex.com, macys.com, gamestop.com, disney.com, kbb.com, accuweather.com} \\
    Website issue & \texttt{nba.com, target.com, phys.org, ups.com, kfc.com, cars.com, bestbuy.com, apartments.com, kohls.com} \\
    \bottomrule
    \end{tabular}}
    \label{tab:web_problem}
\end{table*}

\begin{figure*}[t!]
    \centering
    \begin{subfigure}[t]{0.7\textwidth}
        \includegraphics[width=\linewidth]{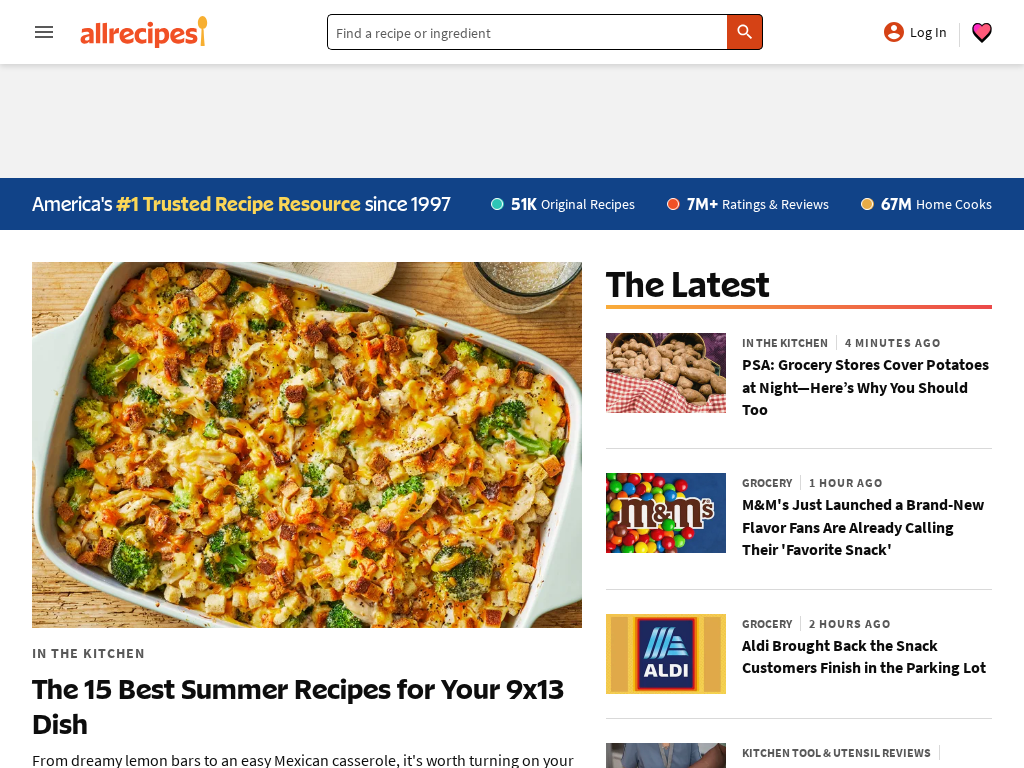}
    \end{subfigure}%
    ~\\
    \begin{subfigure}[t]{\textwidth}
        \centering
        \includegraphics[width=\linewidth]{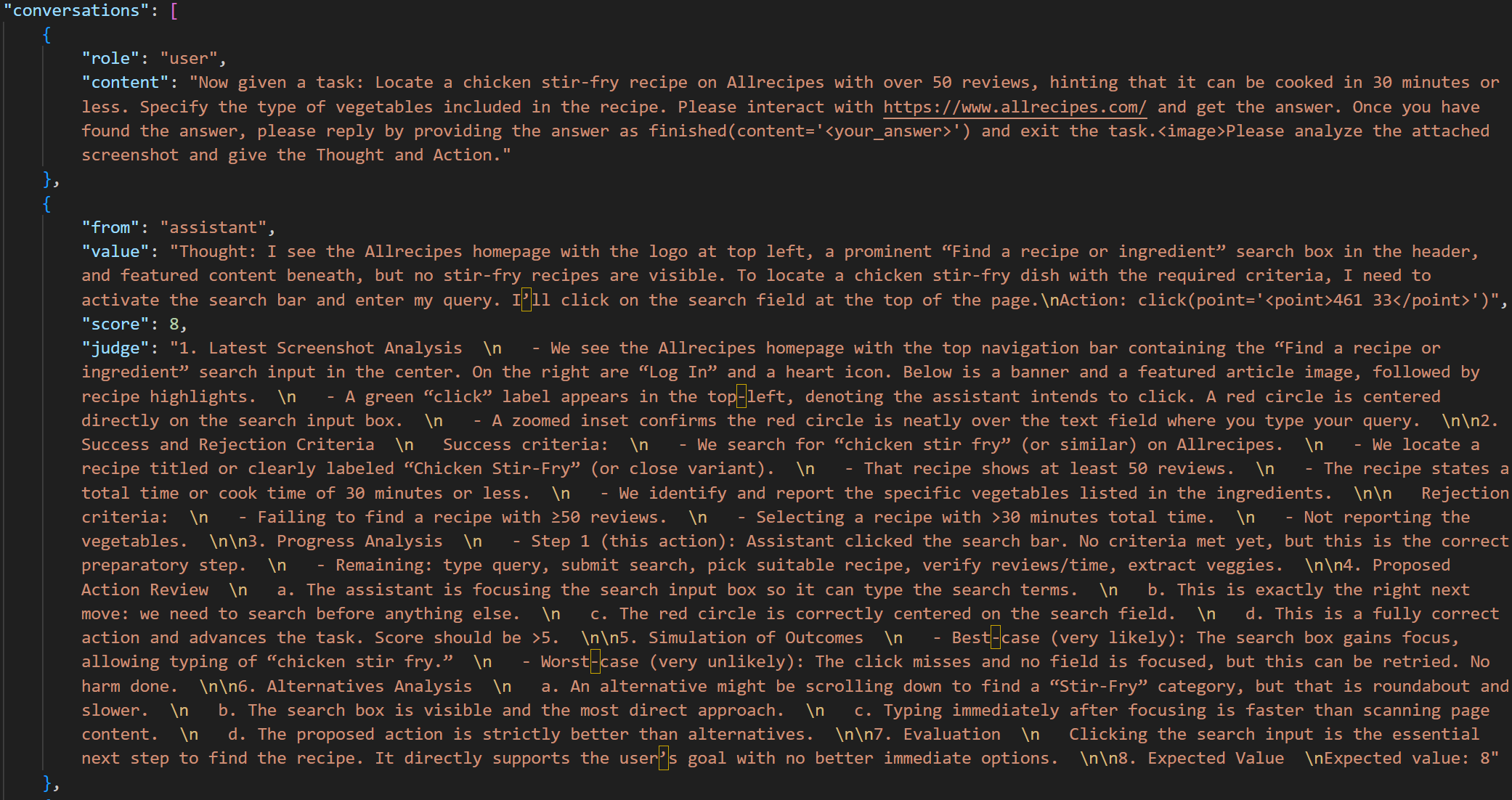}
    \end{subfigure} 
    \caption{Example step for Allrecipe tasks.} 
    \label{fig:example}
\end{figure*}

In \Cref{tab:web_problem}, we list the websites where we have access issues caused by anti-scraping or anti-robot measures as well as bugs within websites. Due to the invalidity, we exclude tasks involving interactions with those websites during training and testing. We also want to clarify that this table only represents the error happening at the time of our testing (August 2025). It is possible that certain tasks are fixed after this time frame, and become valid again. Similarly, we do not include google related tasks in the WebVoyager evaluation, as they have strict human verification requirements.

\section{Data Examples}
An example datapoint in \webstar~is shown in \Cref{fig:example}. Each datapoint contains components: the action from the agent, the thought augmented by \texttt{o4-mini}, the step-level score provided by \texttt{o4-mini}, and the rationale for giving this score. Each step also comes with a screenshot that captures the observation of the current state. Since we do not use any auxiliary source of information such as set of marks, there is no annotation on the screenshot. We have attached more data samples with full trajectories in the attached data folders.

\section{Potential Risks}
This paper presents work whose goal is to advance the field of NLP. There are many potential societal consequences of our work, none which we feel must be specifically highlighted here.

\section{Licenses}
Both WebVoyager and OpenWebVoyager are under Apache-2.0 license. Both Online-Mind2Web Mind2Web-Live are under cc-by-4.0 license.

\section{Prompts for Thought Augmentation and Grading} \label{appendix:prompt}
We attach the full prompts we use for thought augmentation and grading as follows. We use \texttt{o4-mini} to generate the answers for both purposes.

\begin{tcolorbox}[guiPrompt, title={Thought augmentation prompt},
  float*=h,            
  floatplacement=tbp,  
  width=\textwidth
]
\small
You are given the action and thought of the previous several steps of a web browsing agent, the current screenshot annotated with the current action, and a action dictionary describing the specific parameters of the current action. Optionally, you will be given previous screenshots about previous states. The goal is to output the detailed thought process that leads to the current action. 

The action space of the assistant includes:

- click(x, y) where (x, y) are the coordinates of the element to click on.

- scroll(x, y, scroll\_x, scroll\_y) where (x, y) are the coordinates of the element to scroll on and (scroll\_x, scroll\_y) are the scroll amounts in pixels.

- keypress(keys) where keys is a string of keys to press.

- type(text) where text is a string to type. To type in a searchbox, the assistant needs to first click on it, then type.

- wait

- screenshot

- final\_answer(answer) where answer is the final answer to the user task.

Your response should contain the following components:

1. **Situation Description**: Describe your observation of the screenshot in detail. Do not only focus on the regions where the action takes place. Rather, identify key areas and elements that contribute to the decision-making process, such as relevant text, images, or layout features that inform the next steps. Then, arrive at the decision to interact with certain area of interest, and relate it to the goal to achieve. 

2. **Reasoning Alignment**: Ensure your reasoning aligns with the current action and how it contributes to achieving the goal, but avoid using the current action or the annotation on the screenshot as reasoning support, as they represent hindsight rather than predictive insight. Think about the previous steps and how they lead to the current action. Do not output completely equivalent reasoning to the current action even if it is the same action as previous actions. Rather, think about why this action is repeated, and how it is different from previous attempts.

3. **Actionable Instruction**: Conclude with a clear, actionable instruction in one sentence, but no need to use any specific format. Make sure that the instruction matches the provided action.

Important notes:

1. Aim to reason through the task as if solving it, rather than simply reflecting on the outcome. Use the first-person perspective to represent the annotator's thought process.

2. The actions are not necessarily optimal or correct. STICK TO THE GIVEN CURRENT ACTION AND DO NOT COME UP WITH YOUR OWN ACTION or impose your idea about what is the correct action in the current step. Instead, focus on finding rationale and reasoning for the given action.

3. The screenshots are annotated with the actions. On the top left corner, there is the action label, such as click, scroll, wait, type etc. For clicking, there is a red target dot with green label, indicating the clicking position. Do not confuse it with other red elements in the screenshot. For scroll, there is a red target dot for the scrolling centroid, and a red arrow points to the scrolling direction. For drag, there is a red arrow pointing towards the direction of drag, with the dragging start point annotated by a green label.

4. The assistant can only perform one action in each step.

5. If the assistant is on the first step, provide a brief overview of the task and decompose it into smaller steps.

6. The effect of the current action is NOT reflected in the latest screenshot, as the assistant has not executed the action yet. The latest screenshot is just a snapshot of the current state before the action is executed.

7. For final answer, the action is "finished", but the answer shown on the annotated screenshot may be truncated, so focus on the content of the answer provided in text. The agent will finish the task after this step, so do not plan for any further steps or actions.

There is no need to explicitly state the titles of the components, just write them in one paragraph.
\end{tcolorbox}

\begin{tcolorbox}[guiPrompt, title={Grading prompt},
  float*=t,            
  floatplacement=tbp,  
  width=\textwidth
]
\small
You are a critic helping evaluate the next action of a computer use agent. Your goal is to judge the expected value of the proposed next action based on:

1. Whether it meaningfully contributes to successful completion of the user’s task.

2. Whether it is the most promising possible next action from the assistant's available action space — i.e., a "no-regret" choice.

You will be given:

- A USER\_TASK: what the user wants the assistant to do.

- A sequence of prior screenshots (including any zoomed-in crops), each corresponding to one earlier assistant action.

- A single PROPOSED\_NEXT\_ASSISTANT\_ACTION.

- The latest full-screen screenshot and zoomed-in image (if any), annotated with red/green visual guides for where the proposed action is targeting.

------------------------

Assistant action space includes:
- `click(x, y)`: click at coordinates (x, y)

- `scroll(x, y, scroll\_x, scroll\_y)`: scroll at (x, y) by the pixel amounts (scroll\_x, scroll\_y)

- `keypress(keys)`: press keys like "Enter", "Ctrl+A", etc.

- `type(text)`: type a string (must click on an input first)

- `wait`: wait for page to change or update

- `screenshot`: take a new screenshot

- `final\_answer(answer)`: output the final answer to the user (no screenshot will be given with this)

------------------------

SCREENSHOT ANNOTATION CONVENTIONS:
- The top-left green label (e.g. “click”) is the **type of action**, not where it occurs.

- The **red circle** is the exact target position for a click, scroll, or drag — the assistant’s mouse cursor will land there. It must be on the correct element.

- For **scroll actions**, a red arrow shows scroll direction, and red circle marks scroll origin.

- For **drag actions**, red circle marks the start point, red arrow marks direction.

- For sliders, the specific knob dragged and direction both matter. Always examine drag annotations carefully.

IMPORTANT: The assistant has **not yet executed** the proposed action. You must judge its value **before it runs**, based on what is visible on screen.

------------------------

Your task is to follow the 8 steps below, output your analysis for each of the steps, then return an integer score in the format `Expected value: <int>`, where:

- 0 = guaranteed task failure or irreversible error

- 10 = guaranteed task success and no better alternative action exists

- 5 = a borderline step that is either only partially correct or may be outperformed by a better next action

------------------------

1. **Latest Screenshot Analysis**  
   - Then analyze the **latest full screenshot**: describe relevant UI elements visible, current screen state, and the annotation overlays (red circles/arrows, green labels).  
   - Describe any **zoomed-in** image(s). For each, examine where the red circle is at: if it's not directly on an interactive element, say so explicitly (e.g. “not centered on any element”). If a web element (such as a search button) is small, it could be partially obscured by the red circle, pay close attention to such details.
   - If red annotations suggest a different intent than the textual action description, update the interpretation accordingly.  

2. **Success and Rejection Criteria**  
   - Break down the USER\_TASK into **specific, verifiable success conditions**.  
   - Define what would count as incorrect or incomplete (rejection criteria).  

3. **Progress Analysis**  
   - Go through EACH screenshot and earlier action.  
   - For each step, infer what the assistant likely did and how the screen changed.  
   - Mark which success criteria have already been completed, and which remain.

4. **Proposed Action Review**  
   a. Rephrase in plain English what the assistant is trying to do.  
   b. Judge whether this makes sense given the current context and progress.  
   c. If the red circle is off-target or the action does not help, state that the score should be $\leq$5. 
   d. If the action is fully correct and contributes meaningfully to task completion or fixes a past mistake, state that the score should be >5. Specifically, if the current state is already wrong, and the assistant fixes a previous mistake, the score should be higher than 5. EXPLICITLY think about whether the action is fixing a previous mistake.
   Notes for action judgement:
    - For clicking, it does not need to be exactly centered on the element, as long as it is reasonably close.
    - For dragging on sliders, if the knob and direction are correct but the dragging distance is not exact, it can still be considered correct.
    - The assistant cannot type in url, go back to the previous website, or sign in to any website.
    - For final answer, the answer show on the screenshot may be truncated, so focus on the content of the answer provided in text. Do the analysis as other actions. Give score <=5 for final answer only if you are absolutely certain that the answer is incorrect, do not hallucinate about information not provided on the screenshots.

5. **Simulation of Outcomes**  
   a. **Best-case**: if this action executes as intended, what is the best outcome and how likely is it?  
   b. **Worst-case**: if it goes wrong, what’s the worst thing that could happen and how likely is that?

6. **Alternatives Analysis**  
   a. Propose one or more better actions the assistant could take **now**, choosing only from the defined action space.  
   b. Check that these alternatives are viable given what’s visible on screen.  
   c. Rollout likely outcomes for each alternative.  
   d. Compare each alternative to the proposed action — say whether it is **better** or **worse** in terms of task completion.  
   e. If **any alternative** is strictly better, then the proposed action’s score must be $\leq$6. Otherwise, score may be >6.

7. **Evaluation**  
   - Based on all the above, justify the final expected value.  
   - Reiterate whether the action clearly helps, is harmful, partially helpful, or a missed opportunity.  
   - Factor in whether it obeys constraints and sets up a strong next step.

8. **Expected Value**  
   Final output of the value must be on a single line:
Expected value: <int>, where `<int>` is an integer from 0 to 10.
\end{tcolorbox}

\end{document}